\documentclass{article}

\PassOptionsToPackage{numbers}{natbib}
\usepackage[preprint]{any2poster}
\usepackage[utf8]{inputenc} 
\usepackage[T1]{fontenc}    
\usepackage{hyperref}       
\usepackage{url}            
\usepackage{booktabs}       
\usepackage{pifont}         
\usepackage{amsfonts}       
\usepackage{nicefrac}       
\usepackage{microtype}      
\usepackage{xcolor}         
\usepackage{tikz}           
\usepackage{colortbl}
\usepackage{enumitem}
\usepackage{placeins}
\usepackage[most]{tcolorbox}

\newcommand{\cmark}{\ding{51}}
\newcommand{\xmark}{\ding{55}}
\providecommand{\linenumbers}{}
\providecommand{\nolinenumbers}{}

\definecolor{promptbluedark}{RGB}{70,130,180}
\definecolor{promptbg}{RGB}{240,248,255}

\tcbuselibrary{skins, breakable}
\tcbset{
  prompt_func/.style={
    enhanced,
    breakable,
    arc=2pt,
    outer arc=2pt,
    colback=promptbg,
    colframe=promptbluedark,
    boxrule=0.6pt,
    fonttitle=\bfseries\small,
    coltitle=white,
    attach boxed title to top left={yshift=-2mm, xshift=4mm},
    boxed title style={
      colback=promptbluedark,
      colframe=promptbluedark,
      arc=2pt,
      outer arc=2pt,
    },
    left=4pt, right=4pt, top=6pt, bottom=4pt,
  }
}

\definecolor{gt_models}{RGB}{255,214,153}
\definecolor{direct_gen}{RGB}{255,196,117}
\definecolor{multi_agent}{RGB}{255,176,64}
\definecolor{poster_agent}{RGB}{245,128,37}
\newcommand{\ouralg}{Any2Poster Agent}
\newcommand{\ourbench}{Any2Poster Bench}

\title{Any2Poster: Any-Source Poster Generation Across Modalities and Domains}

\author{%
  Amogh Vinaykumar \\
  Flower Mound High School\\
  \texttt{amogh.vinaykumar@gmail.com} \\
  \And
  Aiden Li \\
  University College London\\
  \texttt{yiliu.li@outlook.com} \\
  \And
  Suozhi Huang\thanks{Corresponding authors.} \\
  Princeton University\\
  \texttt{suozhi.huang@princeton.edu} \\
  \And
  Shilong Liu\footnotemark[1] \\
  Princeton University\\
  \texttt{shilong.liu@princeton.edu} \\
}

\begin{document}

\maketitle
\vspace{-1.4em}
{\centering
  \small\href{https://any2poster.github.io/Any2Poster/}{https://any2poster.github.io/Any2Poster/}\par
}
\vspace{0.4em}

\begin{abstract}
Visual posters are a compact medium for communicating dense information, yet progress on automatic poster generation remains difficult to measure because existing evaluations are often restricted to paper-only inputs, narrow domains, or surface-level visual similarity. We introduce \ourbench{}, a benchmark for any-source poster generation that evaluates systems across eight input modalities, including PDFs, URLs, PPTX, DOCX, Markdown, LaTeX, notebooks, and videos, and five content domains. \ourbench{} pairs each source with quiz-based probes of verbatim factual retention and interpretive understanding, together with VLM-based judgments of visual quality, layout, readability, content completeness, and logical flow, enabling reproducible assessment of both information fidelity and visual communication. To instantiate and validate this benchmark, we further present \ouralg{}, an end-to-end reference agent that parses heterogeneous sources, organizes salient content, plans poster layouts, renders posters, and iteratively refines them using visual feedback. On \ourbench{}, \ouralg{} achieves 87.25\% average accuracy across input modalities and 87.28\% across content domains. On PaperQuiz-style evaluation, where prior paper-to-poster agents are directly comparable, \ouralg{} improves over PosterAgent-4o from  51.06–51.33\% to 72.58\% overall accuracy and from 116--121  to 145.16 in density-augmented score. Together, \ourbench{} and \ouralg{} provide a reusable evaluation resource and a competitive baseline for studying multimodal, domain-general poster generation.
\end{abstract}

\section{Introduction}
\label{sec:introduction}

\begin{figure}[t]
    \centering
    \includegraphics[width=0.88\textwidth]{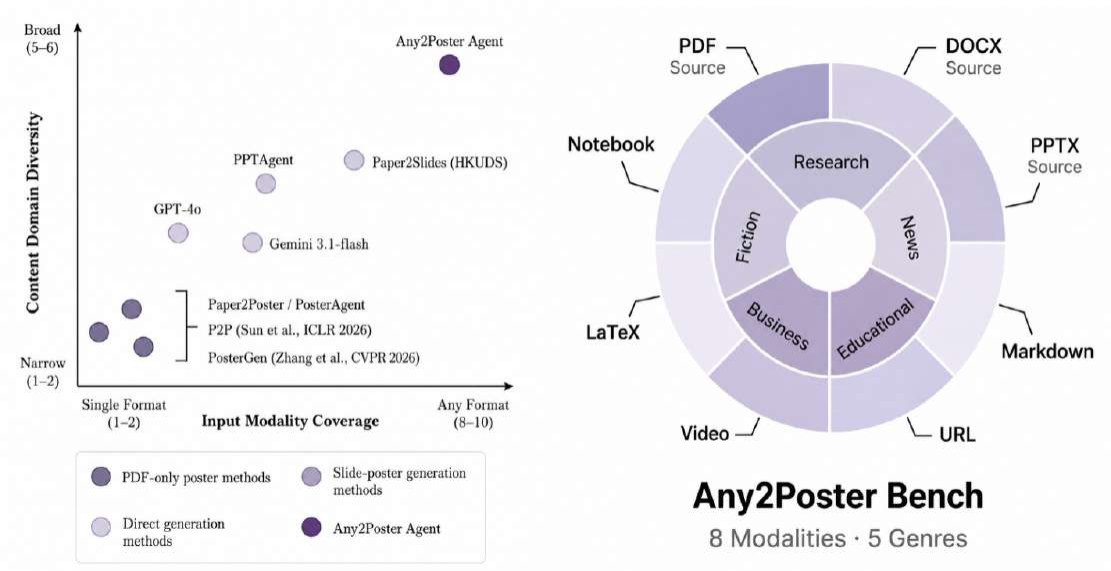}
    \vspace{-0.5em}
    \caption{Positioning of \ourbench{} and representative generation systems
    along input modality coverage and content-domain diversity. Existing
    poster-generation methods mainly focus on single-format academic-paper
    inputs, while \ourbench{} targets broader source modalities and more diverse
    generation scenarios.}
    \label{fig:intro_positioning}
    \vspace{-0.5em}
\end{figure}

Visual posters are widely used for scientific communication, education,
business reporting, and public information sharing. A high-quality poster
compresses rich source content into a single visually coherent page, requiring
content selection, visual hierarchy, spatial planning, and cross-modal
alignment. Prior work shows that even the paper-to-poster setting requires
long-context understanding, multimodal asset extraction, layout planning, and
visual-in-the-loop refinement~\cite{posterbot,paper2poster,p2p,postergen}.

Existing poster-generation evaluations remain limited in scope. Most prior
benchmarks and systems focus on a single input format, especially academic
PDFs. In real-world poster creation, however, users may start from papers,
webpages, slide decks, documents, notebooks, videos, Markdown files, or LaTeX
projects. These sources differ in modality, structure, length, domain, and
communication goal, so a system that works well on papers may fail on weakly
structured prose, business reports, instructional materials, or web-based
content. As shown in Figure~\ref{fig:intro_positioning}, this leaves a gap
between paper-only poster generation and general-purpose poster generation
across modalities and domains.

To address this gap, we introduce \textbf{\ourbench{}}, a benchmark for
\textit{any-source poster generation}. \ourbench{} covers eight input
modalities---PDFs, URLs, PPTX files, DOCX files, Markdown documents, LaTeX
projects, notebooks, and videos---and five content domains: research, news,
education, business, and fiction. It evaluates both cross-modality and
cross-domain generalization, testing whether systems can transform diverse
real-world sources into informative, readable, and visually coherent
single-page posters. To measure information fidelity, \ourbench{} uses
\textbf{BenchQuiz}, a quiz-based reading-comprehension protocol inspired by
PaperQuiz~\cite{paper2poster}, and complements it with VLM-as-judge assessment
of visual and communicative quality.

We also provide \textbf{\ouralg{}} as a strong reference system for
heterogeneous poster generation. \ouralg{} uses unified parsing to convert
heterogeneous inputs into a shared structured representation, performs
content-adaptive poster planning to assign panel roles and visual treatments,
and renders posters through editable HTML/CSS with a lightweight VLM-guided
repair loop for localized visual refinement.

We evaluate \ouralg{} on both \ourbench{} and the existing PaperQuiz-style
paper-to-poster setting. On \ourbench{}, \ouralg{} achieves 87.25\% average
accuracy across eight input modalities and 87.28\% average accuracy across five
content domains, improving over GPT-5 by 3.35 points in cross-modality average
accuracy and over GPT-4o by 11.78 points in cross-domain average accuracy. In
the paper-to-poster setting, \ouralg{} achieves 72.58\% overall accuracy and a
145.16 density-augmented score, improving over PosterAgent-4o from
51.06--51.33\% to 72.58\% overall accuracy and from 116--121 to 145.16 in
density-augmented score.

Our contributions are:
\begin{itemize}[leftmargin=1.2em, itemsep=1pt, topsep=1pt]
    \item We introduce \textbf{\ourbench{}}, a benchmark for any-source poster generation across eight input modalities and five content domains.
    \item We propose \textbf{BenchQuiz}, a quiz-based evaluation protocol for verbatim factual retention and interpretive understanding, complemented by VLM-as-judge visual assessment.
    \item We provide \textbf{\ouralg{}}, a reference system combining unified input parsing, content-adaptive planning, HTML/CSS rendering, and VLM-guided repair.
    \item We show that \ouralg{} generalizes across modalities and domains on \ourbench{} and improves over prior paper-to-poster agents under PaperQuiz-style evaluation.
\end{itemize}
\section{Related Work}
\label{sec:related_work}

Table~\ref{tab:related_work_comparison} compares \ouralg{} with representative
poster and presentation generation systems. Existing systems are typically
designed for paper-only poster generation or slide-deck generation, whereas
\ouralg{} targets heterogeneous inputs with unified parsing, content-adaptive
poster planning, and editable HTML/CSS-based rendering.

\begin{table}[t]
\centering
\vspace{-0.6em}
\scriptsize
\setlength{\tabcolsep}{3pt}
\renewcommand{\arraystretch}{1}
\scalebox{1}{%
\begin{tabular}{lcccc}
\toprule
System
& \shortstack[c]{Beyond\\Paper}
& \shortstack[c]{Unified\\Parsing}
& \shortstack[c]{Adaptive\\Planning}
& \shortstack[c]{HTML/CSS\\Rendering} \\
\midrule
Paper2Poster~\cite{paper2poster}
& \xmark & \xmark & \xmark & \xmark \\
Paper2Slides~\cite{paper2slides}
& \cmark & \xmark & \xmark & \xmark \\
PPTAgent~\cite{pptagent}
& \cmark & \xmark & \xmark & \xmark \\
\rowcolor{poster_agent!15}
\textbf{\ouralg{}}
& \cmark & \cmark & \cmark & \cmark \\
\bottomrule
\end{tabular}%
}
\vspace{0.8em}
\caption{Comparison with representative poster and presentation generation systems.}
\label{tab:related_work_comparison}
\vspace{-0.5em}
\end{table}

\subsection{Poster and Presentation Generation}

Automatic poster generation has been studied in scientific, artistic, and
commercial design settings. Scientific-poster systems such as PosterBot,
Paper2Poster, P2P, PosterGen, PosterSum, and earlier probabilistic approaches
focus on transforming academic papers into posters~\cite{posterbot,
paper2poster,p2p,postergen,postersum,genposter}. Other work studies stylized,
artistic, glyph, product, or controllable poster design~\cite{text2poster2022,
posta,glyphdraw2,planrender,relationdif}. These methods provide useful
insights into poster summarization, layout, and visual aesthetics, but they
typically assume paper-specific inputs or narrower design settings.

Poster generation is also related to document-to-slide and presentation
generation. D2S, SlideGen, multi-stage slide-generation pipelines, persona-aware
slide generation, and PPTAgent study how to convert documents into
presentations~\cite{d2s,slidegen,enhancepre,pre4human,pptagent}. However,
slides distribute information across multiple pages, whereas posters compress
content into a single canvas. This makes poster generation especially sensitive
to information density, spatial hierarchy, local readability, and cross-panel
balance. In contrast to prior settings, \ourbench{} evaluates any-source poster
generation across papers, webpages, slide decks, documents, Markdown files,
LaTeX projects, notebooks, and videos.

\subsection{Document Understanding, Layout, and Rendering}

Any-source poster generation requires robust document understanding and layout
reasoning. Document AI resources and tools such as PubLayNet, DocLayNet,
Nougat, Docling, marker, and BigDocs support layout analysis, document
conversion, and multimodal document understanding~\cite{publaynet,doclaynet,
nougat,docling,marker,bigdocs}. These works motivate the unified parsing stage
of \ouralg{}, which normalizes heterogeneous inputs into a shared structured
representation.

Poster construction also requires visual layout and rendering. LayoutPrompter,
PosterLLaVA, and UI layout generation work study how language or multimodal
models generate structured layouts~\cite{layoutprompter,posterllava,uilayout}.
Design2Code and WebDraw are related to our code-based rendering interface,
where visual artifacts are represented as executable front-end code~\cite{
design2code,webdraw}. Unlike these works, any-source poster generation must
jointly extract source information, summarize it, select or synthesize visuals,
and render an editable single-page poster.

\subsection{Agentic Generation and Evaluation}

LLM-based agents combine reasoning, tool use, code generation, and iterative
repair. ReAct, Toolformer, PAL, ART, and OctoTools show that language models
can benefit from decomposing tasks, calling tools, and executing intermediate
programs~\cite{react,toolformer,gao2022pal,paranjape2023art,octotools}. This
paradigm is well suited to poster generation, where a system must parse
heterogeneous inputs, plan layouts, render editable outputs, inspect visual
artifacts, and repair failures such as overflow or poorly scaled figures.

Evaluation is equally important because a poster must be both visually readable
and faithful to its source. Paper2Poster introduces PaperQuiz and VLM-as-judge
metrics for paper-to-poster evaluation~\cite{paper2poster}. More broadly,
model-based evaluation has been studied in G-Eval, MT-Bench/Chatbot Arena, and
HELM, while VQA and GQA inspire question-answering evaluation for visual
artifacts~\cite{geval,llmjudge,helm,2015vqa,hudson2019gqa}. \ourbench{}
extends these ideas to any-source poster generation by evaluating both
quiz-based information recovery and VLM-as-judge visual communication quality
across multiple modalities and content domains.
\section{Any2Poster Bench}
\label{sec:any2poster_bench}

\begin{figure*}[t]
    \centering
    \includegraphics[width=\textwidth]{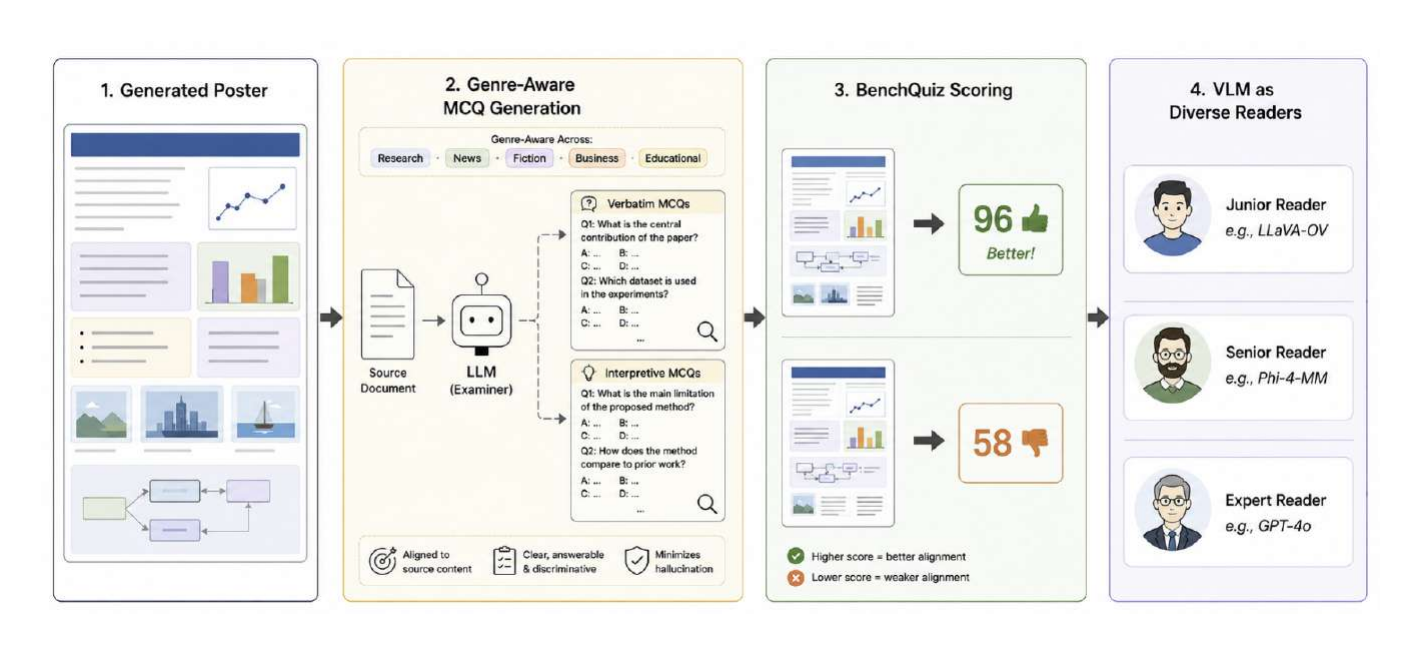}
    \caption{Overview of \textbf{\ourbench{}}. The benchmark covers eight input
    modalities and five content domains, and evaluates generated posters through
    \textit{BenchQuiz}, a genre-aware multiple-choice reading-comprehension
    protocol answered by VLM readers.}
    \label{fig:a2p_bench}
\end{figure*}

We introduce \textbf{\ourbench{}}, a benchmark for evaluating any-source poster
generation across heterogeneous source formats and content domains. As shown in
Figure~\ref{fig:a2p_bench}, the benchmark is designed to test whether a
poster-generation system can handle not only different input modalities, but
also different kinds of source material. The benchmark contains approximately
160 source documents spanning eight modalities: PDF, URL, PPTX, DOCX, Markdown,
LaTeX, notebook, and video. It further covers five content domains: research,
news, educational, business, and fiction. The goal is to move beyond
paper-only poster evaluation and measure whether a system can robustly
transform diverse real-world sources into informative, readable, and visually
coherent single-page posters.

\subsection{Benchmark Construction}
\label{sec:bench_construction}

The benchmark sources are selected according to three principles:
\textit{public accessibility}, \textit{sufficient information density}, and
\textit{diversity in both modality and domain}. Public accessibility ensures
that sources can be documented and, when licensing permits, redistributed or
referenced. Information density ensures that each source contains enough
substantive content to support a meaningful poster and a non-trivial
reading-comprehension evaluation. Modality and domain diversity ensure that the
benchmark tests more than paper-specific parsing or scientific summarization.

Research PDFs and LaTeX projects are drawn from public scientific sources,
while news, educational, business, and fiction sources are collected from
publicly accessible webpages, documents, slide decks, notebooks, and videos.
Video inputs are included only when transcripts are available or can be
reliably extracted. Because naturally occurring Markdown documents that are
both long-form and content-rich are relatively rare outside software
repositories, the Markdown subset is authored specifically for the benchmark to
ensure sufficient information density and domain diversity.

All inputs are stored with a structured manifest that records modality, content
domain, source, license or access information when available, and processing
status. Before a source enters the benchmark, it is manually verified to ensure
that the parser produces valid structured output and that the source contains
enough substantive content to support a full poster. This verification step is
important because any-source poster generation depends not only on visual
generation quality, but also on upstream content extraction and structural
recovery.

\subsection{BenchQuiz Evaluation Protocol}
\label{sec:benchquiz}

To evaluate whether a generated poster preserves and communicates its source
content, we construct \textbf{BenchQuiz}, a genre-aware reading-comprehension
protocol inspired by PaperQuiz~\cite{paper2poster}. For each source, an LLM
examiner generates multiple-choice questions in two categories:
\textit{verbatim} and \textit{interpretive}. Verbatim questions test whether
explicitly stated facts can be recovered from the poster, such as names,
numbers, datasets, methods, events, or claims. Interpretive questions test
whether the poster communicates higher-level meaning, relationships, or
takeaways. In our setup, the examiner generates 20 verbatim and 20
interpretive multiple-choice questions per source.

Questions are answered by VLM readers using only the generated poster. Scores
are computed programmatically and aggregated by modality and by content domain.
We use raw accuracy as the primary BenchQuiz metric because it directly
measures information recoverability from the final poster. For compatibility
with prior paper-to-poster evaluation, we also report density-augmented scoring
when applicable, following the same principle used in Paper2Poster: a poster
should be rewarded for preserving answerable information while remaining
compact~\cite{paper2poster}.

To reduce evaluation bias, answer options are randomly shuffled before each VLM
answering pass so that the correct choice appears uniformly across positions.
VLM readers are instructed to answer \texttt{NA} rather than guess when the
poster does not contain enough evidence for a confident answer. This keeps the
benchmark focused on what the poster actually communicates rather than what a
strong model may infer from prior knowledge. The protocol is related to
visual question-answering evaluation~\cite{2015vqa,hudson2019gqa}, but differs
in that the questions are source-grounded and are designed to test information
transfer through the generated poster.

\subsection{VLM-as-Judge Evaluation}
\label{sec:bench_vlm_judge}

BenchQuiz measures information recoverability, but posters must also be
visually readable and well organized. We therefore complement BenchQuiz with a
VLM-as-judge protocol following recent model-based evaluation practice~\cite{
geval,llmjudge,helm}. The judge scores each poster along six criteria:
element quality, layout balance, engagement, clarity, content completeness,
and logical flow. We average element quality, layout balance, and engagement
as the aesthetic score, and average clarity, content completeness, and logical
flow as the information score. The overall score averages all six dimensions.
Together, these metrics assess both informational effectiveness and visual
communication quality.

\subsection{Dataset Documentation and Intended Use}
\label{sec:bench_documentation}

\ourbench{} is intended for evaluating systems that transform diverse source
materials into single-page visual posters. It is not intended to evaluate
factual correctness beyond the provided source content, nor should it be used
as a general-purpose measure of visual design ability independent of
information fidelity. The benchmark also does not claim to exhaustively cover
all possible poster domains, languages, or visual styles; rather, it provides a
controlled testbed for measuring cross-modality and cross-domain generalization
in poster generation.

Because \ourbench{} is a dataset-centered contribution, dataset documentation
and responsible release are important. Following documentation practices such
as Datasheets for Datasets, Data Cards, and Croissant metadata~\cite{
datasheets,datacards,croissant}, the release includes a manifest documenting
source metadata, modality, content domain, processing status, and evaluation
files. For dataset-centered submission and release, we also prepare
Croissant-compatible metadata and Responsible AI metadata describing intended
uses, limitations, source provenance, licensing considerations when available,
and recommended evaluation practices.
\section{Any2Poster Agent}
\label{sec:any2poster_agent}

\subsection{Framework}
\label{sec:agent_framework}

\begin{figure*}[t]
    \centering
    \includegraphics[width=\textwidth]{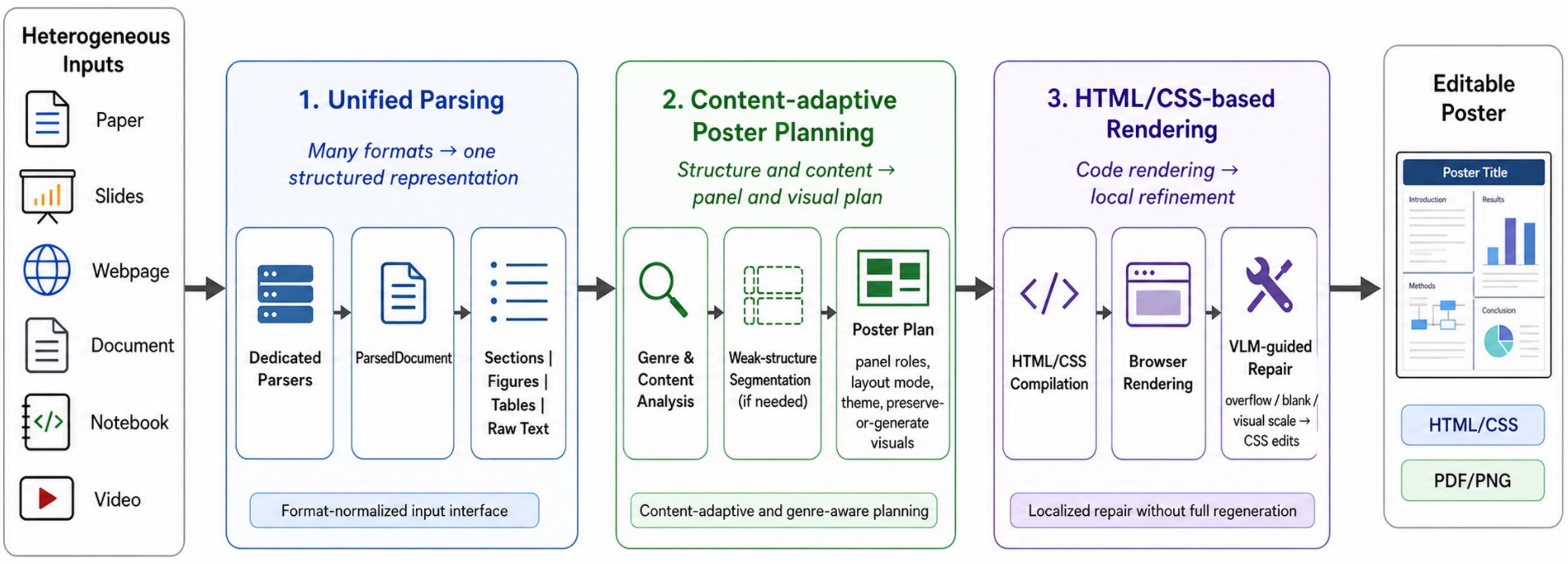}
    \caption{Overview of \textbf{\ouralg{}}. The pipeline has three main
    components: \textbf{(1) Unified Parsing}, which converts heterogeneous
    inputs into a shared structured representation; \textbf{(2) Content-adaptive
    Poster Planning}, which analyzes source content, assigns panel roles and
    layout modes, and decides whether to preserve or generate visuals; and
    \textbf{(3) HTML/CSS-based Rendering}, which compiles an editable poster and
    applies VLM-guided local repair without full regeneration.}
    \label{fig:main_framework}
\end{figure*}

\ouralg{} is a reference system for any-source poster generation. As shown in
Figure~\ref{fig:main_framework}, it converts heterogeneous source inputs into
editable, visually coherent posters through three components:
\textit{unified parsing}, \textit{content-adaptive poster planning}, and
\textit{HTML/CSS-based rendering with VLM-guided repair}. Internally, these
components are instantiated through six operational stages: \textit{parse},
\textit{chunk}, \textit{analyze}, \textit{plan}, \textit{generate}, and
\textit{compile}. The first component normalizes arbitrary inputs into a shared
structured representation. The second component performs content understanding,
weak-structure segmentation when needed, and poster planning conditioned on
content type, content domain, and visual requirements. The third component
renders the poster as editable HTML/CSS, exports browser-rendered outputs, and
iteratively improves panel quality through localized VLM-guided edits.

This design differs from prior poster-generation and presentation-generation
systems in three ways. First, it supports \emph{unified input parsing} rather
than restricting the pipeline to academic papers or a narrow template-retrieval
setting. Second, it performs \emph{content-adaptive planning}, so panel
structure, visual style, and figure usage are selected according to the source
content rather than imposed by a fixed template. Third, it uses
\emph{code as the rendering interface}: poster text is rendered as real
HTML/CSS while visual assets are handled separately, enabling targeted editing
and efficient VLM-guided refinement instead of expensive full-poster
regeneration. This makes \ouralg{} suitable as a strong baseline for
\ourbench{}, where systems must handle diverse input modalities and content
domains.

\subsection{Unified Parsing}
\label{sec:unified_parsing}

The first stage converts each supported input modality into a common
\texttt{ParsedDocument} representation. We use dedicated parsers for different
formats but normalize their outputs into the same schema, consisting of
structured sections, extracted figures with captions, extracted tables, and raw
text. PDF files are parsed with document-conversion tools inspired by recent
work in document layout analysis and structured document understanding~\cite{
publaynet,doclaynet,nougat,docling}. DOCX and PPTX files are handled with
dedicated document parsers. Markdown and plain text are parsed directly into
sectioned text. URLs are processed with boilerplate removal. Video inputs are
converted into transcripts using subtitles or speech transcription. We also
support LaTeX projects and Jupyter notebooks.

The goal of this stage is not merely format conversion, but structural
normalization. Regardless of whether the input is a paper, business report,
story webpage, notebook, or slide deck, downstream stages receive the same type
of object. This unified interface makes later reasoning more robust than
format-specific pipelines, because planning and rendering operate over content
structure instead of ad hoc modality-dependent heuristics.

After parsing, long sections are split into chunks of roughly 800 tokens.
Chunk boundaries prefer paragraph breaks and topic transitions, while equations
and algorithmic blocks are kept intact when possible. Each chunk stores a small
left and right context window so later LLM calls can access local context
without repeatedly processing the full source document.

\subsection{Content-adaptive Poster Planning}
\label{sec:content_adaptive_planning}

The planning stage converts the parsed source into a poster-level content plan.
A global analysis pass predicts the title, core message, section importance,
panel categories, source figures to preserve, and candidate visual suggestions.
A section-level pass then converts each section into panel-ready text and
assigns a layout mode, such as prose only, bullets only, side by side, or
figure dominant. For weakly structured inputs such as fiction or long-form
essays, the agent first synthesizes narrative segments so that continuous prose
does not collapse into a single oversized panel.

The planner also decides whether to preserve source visuals or synthesize new
ones. Extracted figures are retained when they satisfy resolution and
caption-relevance checks; otherwise, the agent generates a visual suggestion
adapted to the content domain. This allows research inputs to preserve
technical diagrams, while business, educational, and narrative inputs can use
more explanatory or story-oriented visuals.

\subsection{HTML/CSS-based Rendering}
\label{sec:html_css_rendering}

The generation and compilation stages use code, rather than an image model, as
the primary interface for poster construction. All poster text is rendered in
real HTML/CSS, which gives character-level fidelity that image-based text
rendering often cannot guarantee. Each panel is instantiated as a structured
HTML block whose layout is determined by the planning stage. Visuals are either
preserved source figures or newly generated assets. For generated visuals, the
system constructs prompts from the panel concept, visual type, validated
numeric data, and active design theme.

This code-based interface is related to recent design-to-code and automatic
layout-generation work~\cite{layoutprompter,posterllava,design2code,uilayout}.
In our setting, code-based rendering gives the agent fine-grained control over
text, layout, and visual assets. Since these components are separated, the
system can edit typography, spacing, and figure sizing directly through code
without discarding the rest of the poster. The renderer embeds figures as data
URIs, assembles a complete HTML poster, and uses a browser renderer to export
both PDF and PNG outputs at poster resolution.

We also add a lightweight validation step for data-dependent visuals such as
bar charts, line charts, comparisons, and tables. For these cases, the model is
asked to return only data points explicitly supported by the source text;
unsupported values are discarded before image generation. This reduces
hallucinated quantitative content and improves the trustworthiness of
synthesized visuals.

\subsection{Visual Feedback Loop with VLMs}
\label{sec:visual_feedback_loop}

After rendering, \ouralg{} applies an optional VLM-based feedback loop for
localized visual repair. The agent crops each poster panel, asks a VLM to
detect issues such as overflow, excessive blank space, or poorly scaled
visuals, and maps the diagnosis to targeted HTML/CSS edits before re-rendering.
Unlike prior systems that rely on broader panel regeneration~\cite{paper2poster},
this mechanism preserves already correct content while improving readability
and layout quality. It also differs from presentation-generation systems that
lack a comparable single-page poster repair loop~\cite{pptagent,paper2slides}.
\section{Experiments}
\label{sec:experiments}

We evaluate \ouralg{} on both \ourbench{} and the existing PaperQuiz-style
paper-to-poster setting. The experiments address four questions: (i) whether
\ouralg{} generalizes across heterogeneous input modalities, (ii) whether it
generalizes across content domains, (iii) whether it remains competitive with
prior paper-to-poster agents, and (iv) whether the generated posters preserve
both recoverable information and visual communication quality. We use
quiz-based accuracy as the primary measure of information recovery and
complement it with VLM-as-judge assessment.

\subsection{Experimental Setup}
\label{sec:exp_setup}

\paragraph{Any2Poster Bench evaluation.}
\ourbench{} evaluates any-source poster generation across eight input
modalities: PDF, URL, PPTX, DOCX, Markdown, LaTeX, notebook, and video. For
each evaluation instance, we generate 20 verbatim and 20 interpretive
multiple-choice questions and measure the fraction answered correctly from the
generated poster. Verbatim questions test whether factual details such as
titles, datasets, numbers, and named methods are explicitly recoverable from
the poster. Interpretive questions test whether the poster communicates
higher-level understanding, such as the source's main contribution,
conclusions, and implications. We aggregate accuracy by input modality and by
content domain.

\paragraph{Baselines.}
We compare \ouralg{} with general-purpose multimodal models, including
GPT-4o, Gemini 2.5 Flash, and GPT-5~\cite{gpt4o,gemini25,gpt5}. We also
compare with Paper2Poster when the input format is supported~\cite{
paper2poster}. Since Paper2Poster is designed for paper-like inputs, it is
reported as \textsc{N/A} on unsupported modalities. For PaperQuiz-style
evaluation, we compare against Paper2Poster-Qwen and PosterAgent-4o following
the prior paper-to-poster setting~\cite{paper2poster,qwen25vl}.

\paragraph{VLM-as-judge evaluation.}
In addition to quiz-based information recovery, we use VLM-as-judge scoring to
assess visual and communicative quality. The judge scores each poster along six
criteria: element quality, layout balance, engagement, clarity, content
completeness, and logical flow. We additionally report an aesthetic score,
computed from element quality, layout balance, and engagement; an information
score, computed from clarity, content completeness, and logical flow; and an
overall score averaging all six dimensions.

\subsection{Cross-Modality Evaluation}
\label{sec:cross_modality}

A central goal of \ourbench{} is to evaluate whether a poster-generation system
can handle heterogeneous source formats rather than only scientific papers.
Table~\ref{tab:modality_results} reports quiz accuracy across eight input
modalities. \ouralg{} achieves the best average accuracy, 87.25\%,
outperforming GPT-4o, Gemini 2.5 Flash, GPT-5, and Paper2Poster. The strongest
results are obtained on Markdown and LaTeX inputs, both above 93\%. PPTX
remains the most challenging modality for \ouralg{} at 75.00\%, suggesting that
slide inputs require further layout- and structure-aware parsing. Although
GPT-5 performs slightly better on PDF and PPTX, \ouralg{} is more robust on
average across the full modality spectrum.

\begin{table*}[t]
\centering
\scriptsize
\setlength{\tabcolsep}{3.2pt}
\caption{Cross-modality quiz accuracy (\%) on \ourbench{}. Each modality is
evaluated using verbatim and interpretive questions. \textsc{N/A} indicates
that the method does not support the corresponding input modality.}
\label{tab:modality_results}
\begin{tabular}{lccccccccc}
\toprule
\textbf{Method} &
\textbf{PDF} &
\textbf{URL} &
\textbf{PPTX} &
\textbf{DOCX} &
\textbf{MD} &
\textbf{LaTeX} &
\textbf{Notebook} &
\textbf{Video} &
\textbf{Avg.} \\
\midrule
GPT-4o &
75.50 & 72.50 & 74.00 & 75.00 & 72.50 & 80.00 & 62.50 & 80.00 & 74.00 \\
Gemini 2.5 Flash &
72.50 & 71.20 & 73.80 & 73.80 & 80.00 & 80.00 & 79.00 & 78.80 & 76.14 \\
GPT-5 &
\textbf{86.20} & 82.50 & \textbf{78.80} & 82.50 & 85.20 & 83.70 & 85.00 & 87.40 & 83.90 \\
Paper2Poster &
57.60 & \textsc{N/A} & \textsc{N/A} & \textsc{N/A} & \textsc{N/A} & 59.00 & \textsc{N/A} & \textsc{N/A} & 58.30 \\
\ouralg{} &
85.63 & \textbf{88.13} & 75.00 & \textbf{86.88} &
\textbf{93.00} & \textbf{93.13} & \textbf{86.88} & \textbf{89.38} &
\textbf{87.25} \\
\bottomrule
\end{tabular}
\end{table*}

\subsection{Cross-Domain Evaluation}
\label{sec:cross_domain}

We next evaluate whether \ouralg{} generalizes across content domains.
Table~\ref{tab:domain_results} reports quiz accuracy across fiction, research,
business, educational, and news sources. \ouralg{} achieves an average accuracy
of 87.28\%, compared with 75.50\% for GPT-4o and 68.50\% for Paper2Poster on
its supported domains. Performance is consistent across domains, with all
domain-level accuracies above 86\%. These results show that \ourbench{}
evaluates a broader capability than scientific paper summarization alone:
systems must extract, organize, and visually communicate information from
sources with different discourse structures.

\begin{table}[t]
\centering
\scriptsize
\setlength{\tabcolsep}{4pt}
\caption{Cross-domain quiz accuracy (\%) on \ourbench{}. Higher is better.
$^\dagger$Paper2Poster average is computed over supported domains only because
the educational setting is unsupported.}
\label{tab:domain_results}
\begin{tabular}{lcccccc}
\toprule
\textbf{Method} &
\textbf{Fiction} &
\textbf{Research} &
\textbf{Business} &
\textbf{Educational} &
\textbf{News} &
\textbf{Avg.} \\
\midrule
GPT-4o &
80.00 & 72.50 & 82.50 & 62.50 & 80.00 & 75.50 \\
Paper2Poster &
61.00 & 74.00 & 68.00 & \textsc{N/A} & 71.00 & 68.50$^\dagger$ \\
\ouralg{} &
\textbf{89.38} & \textbf{86.57} & \textbf{87.15} &
\textbf{86.79} & \textbf{86.50} & \textbf{87.28} \\
\bottomrule
\end{tabular}
\end{table}

\subsection{Comparison with Prior Paper-to-Poster Agents}
\label{sec:paperquiz_results}

Table~\ref{tab:paperquiz} compares \ouralg{} with prior paper-to-poster agent
baselines under PaperQuiz-style evaluation. \ouralg{} achieves substantially
higher overall accuracy than both Paper2Poster-Qwen and PosterAgent-4o. In
particular, it improves over PosterAgent-4o from 51.06--51.33\% to 72.58\%
overall accuracy and from 116--121 to 145.16 in density-augmented score. The
gain is especially large on interpretive questions, increasing from 49.17\%
for PosterAgent-4o to 84.00\% for \ouralg{}. This suggests that \ouralg{}
better communicates the high-level meaning of the source rather than only
preserving isolated facts.

\begin{table}[t]
\centering
\scriptsize
\setlength{\tabcolsep}{4pt}
\caption{PaperQuiz-style comparison with prior paper-to-poster agents.
Verbatim, interpretive, and overall values are accuracies in percentage points.
Higher is better for all metrics.}
\label{tab:paperquiz}
\begin{tabular}{lcccc}
\toprule
\textbf{Method} &
\textbf{Verbatim} &
\textbf{Interpretive} &
\textbf{Overall Acc.} &
\textbf{Density-Aug.} \\
\midrule
Paper2Poster-Qwen &
51.81 &
48.79 &
50.30--50.57 &
114--115 \\
PosterAgent-4o &
52.95 &
49.17 &
51.06--51.33 &
116--121 \\
\ouralg{} &
\textbf{60.80} &
\textbf{84.00} &
\textbf{72.58} &
\textbf{145.16} \\
\bottomrule
\end{tabular}
\end{table}

\subsection{VLM-as-Judge Results}
\label{sec:vlm_judge_results}

Table~\ref{tab:vlm_judge} reports VLM-as-judge results. Compared with
Paper2Poster, \ouralg{} improves the overall VLM-as-judge score from 3.69 to
4.03. The largest gain appears in logical flow, which increases from 3.64 to
4.96, indicating that the generated poster better organizes information into a
coherent visual narrative. \ouralg{} also improves element quality, layout
balance, engagement, clarity, content completeness, aesthetic score, and
information score.

\begin{table*}[t]
\centering
\scriptsize
\setlength{\tabcolsep}{4pt}
\caption{VLM-as-judge evaluation. Element, layout, and engagement form the
aesthetic score; clarity, content completeness, and logical flow form the
information score. Higher is better.}
\label{tab:vlm_judge}
\vspace{0.7em}
\begin{tabular}{lccccccccc}
\toprule
\textbf{Method} &
\textbf{Elem.} &
\textbf{Layout} &
\textbf{Engage.} &
\textbf{Clarity} &
\textbf{Content} &
\textbf{Flow} &
\textbf{Aesthetic} &
\textbf{Info.} &
\textbf{Overall} \\
\midrule
Paper2Poster &
3.94 & 3.77 & 2.91 & 3.99 & 3.91 & 3.64 & 3.54 & 3.85 & 3.69 \\
\ouralg{} &
\textbf{4.04} & \textbf{3.93} & \textbf{3.13} & \textbf{4.04} &
\textbf{4.07} & \textbf{4.96} & \textbf{3.70} & \textbf{4.36} &
\textbf{4.03} \\
\bottomrule
\end{tabular}
\end{table*}

\subsection{Ablation Study}
\label{sec:ablation}

We ablate two key components of \ouralg{}: unified input parsing and visual
feedback refinement. Removing unified input parsing reduces verbatim accuracy
from 60.80\% to 31.00\%, showing that structure-aware parsing is critical for
preserving fine-grained source facts. Removing visual feedback refinement
reduces overall accuracy from 72.58\% to 55.50\%, indicating that visual
quality control directly affects whether downstream readers can recover both
factual and interpretive information. The full ablation table is provided in
Appendix~\ref{app:ablation}.

We report qualitative examples and token consumption in
Appendices~\ref{app:qualitative} and~\ref{app:efficiency}. Overall, the
results show that \ourbench{} reveals capabilities not captured by paper-only
benchmarks, and that \ouralg{} provides a strong reference system for
any-source poster generation.

\FloatBarrier
\section{Conclusion}
\label{sec:conclusion}

We introduced \ourbench{}, a benchmark for evaluating any-source poster
generation across eight input modalities and five content domains. By combining
BenchQuiz with VLM-as-judge assessment, \ourbench{} measures both information
recoverability and visual communication quality. We further provided
\ouralg{} as a reference system that combines unified parsing, content-adaptive
planning, HTML/CSS rendering, and VLM-guided repair. Experiments show that
\ouralg{} generalizes across modalities and domains on \ourbench{} and improves
over prior paper-to-poster agents under PaperQuiz-style evaluation. We hope
\ourbench{} supports future work on reliable, multimodal, and domain-general
visual communication systems.

\bibliography{references}
\bibliographystyle{plain} 


\clearpage
\appendix
\nolinenumbers
\section*{Appendix}
\addcontentsline{toc}{section}{Appendix}

\noindent\textbf{Contents}

\vspace{1em}

\newlength{\tocindentone}\setlength{\tocindentone}{1.2em}
\newlength{\tocindentwo}\setlength{\tocindentwo}{2.6em}

\noindent\textcolor{red}{\textbf{A}}\hspace{0.6em}Dataset Details%
  \leaders\hbox{.}\hfill\textbf{\pageref{sec:dataset-details}}\\[6pt]

\noindent\textcolor{red}{\textbf{B}}\hspace{0.6em}Prompt Templates%
  \leaders\hbox{.}\hfill\textbf{\pageref{sec:prompts}}\\[6pt]

\noindent\hspace{\tocindentone}B.1\hspace{0.5em}Analysis Prompts%
  \leaders\hbox{.}\hfill\pageref{sec:prompts-analysis}\\[6pt]

\noindent\hspace{\tocindentone}B.2\hspace{0.5em}Generation Prompts%
  \leaders\hbox{.}\hfill\pageref{sec:prompts-generation}\\[6pt]

\noindent\hspace{\tocindentone}B.3\hspace{0.5em}Feedback Prompts%
  \leaders\hbox{.}\hfill\pageref{sec:prompts-feedback}\\[6pt]

\noindent\hspace{\tocindentone}B.4\hspace{0.5em}Evaluation Prompts%
  \leaders\hbox{.}\hfill\pageref{sec:prompts-eval}\\[6pt]

\noindent\hspace{\tocindentwo}B.4.1\hspace{0.5em}PaperQuiz: Question Generation%
  \leaders\hbox{.}\hfill\pageref{sec:prompts-paperquiz-gen}\\[6pt]

\noindent\hspace{\tocindentwo}B.4.2\hspace{0.5em}PaperQuiz: Poster Answering%
  \leaders\hbox{.}\hfill\pageref{sec:prompts-paperquiz-answer}\\[6pt]

\noindent\hspace{\tocindentwo}B.4.3\hspace{0.5em}BenchQuiz: Genre-Aware Evaluation%
  \leaders\hbox{.}\hfill\pageref{sec:prompts-benchquiz}\\[6pt]

\noindent\hspace{\tocindentwo}B.4.4\hspace{0.5em}VLM-as-Judge%
  \leaders\hbox{.}\hfill\pageref{sec:prompts-vlm-judge}\\[6pt]

\noindent\textcolor{red}{\textbf{C}}\hspace{0.6em}Implementation Details%
  \leaders\hbox{.}\hfill\textbf{\pageref{sec:impl}}\\[6pt]

\noindent\textcolor{red}{\textbf{D}}\hspace{0.6em}Ablation Study%
  \leaders\hbox{.}\hfill\textbf{\pageref{app:ablation}}\\[6pt]

\noindent\textcolor{red}{\textbf{E}}\hspace{0.6em}Efficiency%
  \leaders\hbox{.}\hfill\textbf{\pageref{sec:efficiency}}\\[6pt]

\noindent\textcolor{red}{\textbf{F}}\hspace{0.6em}Additional Qualitative Examples%
  \leaders\hbox{.}\hfill\textbf{\pageref{sec:qualitative}}\\[6pt]

\noindent\textcolor{red}{\textbf{G}}\hspace{0.6em}Limitations%
  \leaders\hbox{.}\hfill\textbf{\pageref{sec:limitations}}\\[6pt]

\noindent\textcolor{red}{\textbf{H}}\hspace{0.6em}Responsible Release%
  \leaders\hbox{.}\hfill\textbf{\pageref{sec:responsible}}\\[6pt]

\clearpage
\linenumbers

\section{Dataset Details}
\label{app:dataset_details}\label{sec:dataset-details}

\subsection{Benchmark Composition}
\label{app:benchmark_composition}

\ourbench{} is designed to evaluate any-source poster generation across both input modality and content domain. The benchmark contains approximately 160 source documents spanning eight input modalities---PDF, URL, PPTX, DOCX, Markdown, LaTeX, notebook, and video---and five content domains: research, news, educational, business, and fiction.

Of these, 32 fully evaluated instances are released here, with one instance for each modality--domain cell, covering all eight modalities and all five domains. The remaining instances are withheld because their source licenses do not permit redistribution of extracted text, such as news articles, institutional presentations, and proprietary reports. Poster outputs and evaluation scores for those withheld instances will be included in the full public release.

Source attribution and URLs for all 32 released instances are provided in \texttt{benchmark\_manifest.csv}. Each source is paired with BenchQuiz questions and metadata describing its modality, content domain, source type, processing status, and associated evaluation files.

The benchmark construction follows three principles. First, sources should be
publicly accessible or otherwise documentable. Second, each source should
contain enough information density to support a meaningful single-page poster.
Third, the collection should cover heterogeneous structures, including
well-structured scientific papers, semi-structured documents and slide decks,
webpage content, executable notebooks, video transcripts, and weakly structured
long-form prose. This diversity is intended to test whether a poster-generation
system can generalize beyond paper-only inputs.

\subsection{Source Processing}
\label{app:source_processing}

Each source is converted into a structured representation before poster
generation. For document-like inputs, the parser extracts section text, figures,
captions, tables, and raw text when available. For URL inputs, boilerplate
content is removed before extraction. For video inputs, transcripts are obtained
from subtitles or speech transcription when available. For notebooks, markdown
cells, code cells, outputs, and figures are converted into a document-like
representation. For LaTeX inputs, source files are parsed into sectioned text
and associated assets when possible.

Before inclusion in the benchmark, each source is manually checked to ensure
that the parser produces valid structured output and that the source contains
enough substantive content to support poster generation and question-answering
evaluation. Sources that fail parsing or contain insufficient content are
excluded or replaced.

\section{Prompt Templates}
\label{sec:prompts}

We present the full prompt templates used by \textsc{Any2Poster} Agent across its
pipeline stages: \textbf{Analysis} (global document understanding and per-section
distillation), \textbf{Generation} (title banner, panel rendering, and visual
synthesis), \textbf{Feedback} (panel layout review), and \textbf{Evaluation}
(PaperQuiz question generation and answering, BenchQuiz, and VLM-as-Judge).

\subsection{Analysis Prompts}
\label{sec:prompts-analysis}

The analysis stage runs two LLM passes: a global pass that extracts
poster-level metadata and visual suggestions, and a per-section pass that
distills each section into poster-ready content. For headerless documents
(e.g.\ fiction, news), a prose-segmentation prompt is used instead.

\textbf{Global Document Analyzer.}

\begin{tcolorbox}[prompt_func, title={Prompt: Global Document Analyzer}]
\textbf{System Prompt:} You are an expert information poster designer who creates
visual summaries for any type of content---academic research papers, news
articles, business reports, educational materials, and creative or narrative
works. You analyze documents and produce structured JSON that guides poster
creation. You identify the core message, the single most impactful element, and
determine which sections and visuals best represent the content on a poster.
You MUST respond with valid JSON only. Never refuse to process a document---
every document has a story worth visualizing.

\medskip
\textbf{Instructions:} Analyze this document for poster creation. Respond with
JSON fields including \texttt{poster\_title}, \texttt{authors},
\texttt{affiliation}, \texttt{key\_contribution}, \texttt{headline\_result},
\texttt{summary}, \texttt{narrative\_arc}, \texttt{paper\_domain},
\texttt{methodology\_summary}, \texttt{results\_summary},
\texttt{suggested\_color\_theme}, \texttt{venue},
\texttt{sections\_to\_include}, \texttt{section\_importance},
\texttt{section\_categories}, \texttt{essential\_figure\_ids}, and
\texttt{visual\_suggestions}.

\medskip
\textbf{Rules:}
\begin{itemize}[leftmargin=1.2em, itemsep=1pt, topsep=1pt]
  \item Include 7--9 sections and never more than 9.
  \item Each section must cover a distinct aspect.
  \item Skip references, acknowledgments, and pure metadata sections.
  \item Use the headline result as the single most impactful element.
  \item Suggest 6--8 visuals and at least three visual types.
\end{itemize}
\end{tcolorbox}

\textbf{Per-Section Content Distiller.}

\begin{tcolorbox}[prompt_func, title={Prompt: Section Content Distiller}]
\textbf{System Prompt:} You are an expert at distilling any document into
poster-ready content. You decide whether each section is best presented as
flowing prose, structured bullets, or a mix. Each bullet must be self-contained,
specific with facts, names, quotes, or data, and under 10 words. You MUST
respond with valid JSON only.

\medskip
\textbf{Instructions:} Extract poster-ready content from this section using the
global poster title, key contribution, headline result, domain or genre, section
importance, and section category.

Respond with \texttt{poster\_section\_title}, \texttt{content\_type},
\texttt{lead\_paragraph}, \texttt{bullets}, \texttt{sub\_headers},
\texttt{key\_message}, \texttt{provenance}, and
\texttt{recommended\_figure\_ids}.

\medskip
\textbf{Rules:} Use prose for background and conclusions, bullets for results
and key findings, and mixed format for methodology or architecture. Each bullet
must contain a specific fact and convey unique information.
\end{tcolorbox}

\textbf{Prose Segment Extractor (Headerless Documents).}

\begin{tcolorbox}[prompt_func, title={Prompt: Prose Segment Extractor}]
\textbf{System Prompt:} You are a document analyst. Identify the main thematic
or narrative segments of the given document and extract the most representative
passage from each. You MUST respond with valid JSON only.

\medskip
\textbf{Instructions:} Identify exactly 8 distinct thematic, narrative, or
topical segments that would make excellent poster panels. For each segment,
provide a short, evocative title and the most representative 200--350 words
extracted verbatim from the document.

\medskip
Required segment structure: opening or inciting incident; characters and
setting; central conflict; key scene; deepening complications; climax;
resolution; and themes and significance.
\end{tcolorbox}

\subsection{Generation Prompts}
\label{sec:prompts-generation}

\textbf{Poster Title Banner.}

\begin{tcolorbox}[prompt_func, title={Prompt: Poster Title Banner}]
Generate an academic conference poster title banner in an ultra-wide landscape
format. Render the title, authors, affiliation, and headline result when
present.

\medskip
\textbf{Layout Rules:}
\begin{itemize}[leftmargin=1.2em, itemsep=1pt, topsep=1pt]
  \item Full-width poster banner, not a small title card.
  \item Bold title font, clearly larger than authors.
  \item Authors line is mandatory and directly below the title.
  \item Keep text compact with safe padding and no clipping.
\end{itemize}
\end{tcolorbox}

\textbf{Poster Panel.}

\begin{tcolorbox}[prompt_func, title={Prompt: Poster Panel}]
Generate a single panel for an academic research poster with a clean,
professional style. Use a flat white background panel with a header bar at the
top containing the section title in white text.

\medskip
\textbf{Layout Rules:} Use clean sans-serif typography, consistent spacing,
large readable titles, no overlap, no decorative ornaments, and enough reserved
space for a visual when one is placed below the text.
\end{tcolorbox}

\textbf{Visual Generation.}

\begin{tcolorbox}[prompt_func, title={Prompt: Visual Generation}]
Generate a professional visual for an information poster using the supplied
concept, visual type, and content domain.

\medskip
\textbf{Style Requirements:}
\begin{itemize}[leftmargin=1.2em, itemsep=1pt, topsep=1pt]
  \item Use a clean vector style and avoid photographic stock imagery.
  \item Use readable sans-serif labels and high contrast.
  \item Match the poster panel background exactly.
  \item Make the visual specific to the document's actual content.
\end{itemize}
\end{tcolorbox}

\subsection{Feedback Prompts}
\label{sec:prompts-feedback}

After an initial render, a VLM feedback loop inspects each panel crop and
returns a structured issue code. Deterministic HTML/CSS mutations are then
applied to address detected problems.

\begin{tcolorbox}[prompt_func, title={Prompt: Panel Layout Reviewer}]
\textbf{System Prompt:} You are a strict poster layout reviewer. Only report
real, visible issues. You MUST respond with valid JSON and use only allowed
issue codes.

\medskip
Allowed issue codes: \texttt{good}, \texttt{overflow}, \texttt{too\_blank},
\texttt{visual\_too\_small}, and \texttt{visual\_too\_large}. If there are no
issues, return \texttt{\{"issues": ["good"]\}}.
\end{tcolorbox}

\subsection{Evaluation Prompts}
\label{sec:prompts-eval}

We adopt the PaperQuiz and VLM-as-Judge evaluation protocols from
Paper2Poster~\citep{paper2poster} for comparability. We additionally
introduce BenchQuiz for evaluating non-research documents.

\subsubsection{PaperQuiz: Question Generation}
\label{sec:prompts-paperquiz-gen}

\begin{tcolorbox}[prompt_func, title={Prompt: Generate Verbatim QA (PaperQuiz)}]
Read the supplied Markdown text and produce exactly 50 multiple-choice QA items
whose answers can be located verbatim or nearly verbatim in that text. Questions
must be suitable for conference-poster readers and must exclude references,
citations, author acknowledgements, and citation minutiae. Output only a JSON
object with balanced answer choices.
\end{tcolorbox}

\begin{tcolorbox}[prompt_func, title={Prompt: Generate Interpretive QA (PaperQuiz)}]
Read the supplied Markdown text and create exactly 50 multiple-choice questions
that capture high-level understanding of the work, including purpose, novelty,
core approach, findings, implications, limitations, and conclusions. Output only
the final JSON object.
\end{tcolorbox}

\subsubsection{PaperQuiz: Poster Answering}
\label{sec:prompts-paperquiz-answer}

\begin{tcolorbox}[prompt_func, title={Prompt: Answer Questions (PaperQuiz)}]
Answer each multiple-choice question based solely on the poster image. If there
is sufficient evidence, choose the option and include a brief supporting poster
region. If the poster does not contain enough information, respond with
\texttt{NA} for both answer and reference. Output only the required JSON object.
\end{tcolorbox}

\subsubsection{BenchQuiz: Genre-Aware Evaluation}
\label{sec:prompts-benchquiz}

For non-research documents, BenchQuiz is a genre-aware variant with 20 questions
per document: 10 verbatim and 10 interpretive. A \texttt{genre\_context} field is
resolved at runtime.

\begin{tcolorbox}[prompt_func, title={Prompt: Generate Verbatim QA (BenchQuiz)}]
Read the supplied source document and produce exactly 20 multiple-choice
questions that test retention of concrete details. Every answer must be grounded
in an explicit phrase, statistic, name, date, or direct statement in the source.
Output only the JSON object and balance answers across options.
\end{tcolorbox}

\begin{tcolorbox}[prompt_func, title={Prompt: Generate Interpretive QA (BenchQuiz)}]
Read the supplied source document and produce exactly 20 multiple-choice
questions that test high-level understanding: central argument, narrative,
purpose, significance, and primary takeaways. Output only the JSON object.
\end{tcolorbox}

\subsubsection{VLM-as-Judge}
\label{sec:prompts-vlm-judge}

Each poster is scored on six criteria by a VLM judge. For each criterion, the
model returns \texttt{\{"reason": "...", "score": <1--5>\}}. We use the
same rubrics as Paper2Poster~\citep{paper2poster}.

\begin{tcolorbox}[prompt_func, title={Prompt: Element Quality Judge}]
Judge the clarity, consistency, resolution, labeling, legends, and visual style
of figures, charts, and images. Use a conservative 1--5 scale and return a JSON
object with a short reason and integer score.
\end{tcolorbox}

\begin{tcolorbox}[prompt_func, title={Prompt: Layout Balance Judge}]
Judge the arrangement of text blocks, headings, figures, whitespace, alignment,
and reading path. Penalize overlap, crowding, inconsistent alignment, and weak
hierarchy. Return a JSON reason and integer score.
\end{tcolorbox}

\begin{tcolorbox}[prompt_func, title={Prompt: Engagement Judge}]
Judge color harmony, typography, visual balance, and the poster's ability to
grab and hold attention. Reserve high scores for exemplary work. Return a JSON
reason and integer score.
\end{tcolorbox}

\begin{tcolorbox}[prompt_func, title={Prompt: Clarity Judge}]
Judge sentence-level clarity, grammar, phrasing, terminology, and intra-section
coherence. Penalize awkward phrasing, jargon, and grammatical issues. Return a
JSON reason and integer score.
\end{tcolorbox}

\begin{tcolorbox}[prompt_func, title={Prompt: Content Completeness Judge}]
Judge whether the poster includes all essential sections with sufficient detail,
including objectives, methods, results, interpretation, and limitations. Return a
JSON reason and integer score.
\end{tcolorbox}

\begin{tcolorbox}[prompt_func, title={Prompt: Logical Flow Judge}]
Judge whether the major sections connect into a coherent narrative. Penalize
missing transitions, weak logical links, and disjointed structure. Return a JSON
reason and integer score.
\end{tcolorbox}

\section{Implementation Details}
\label{app:implementation}\label{sec:impl}

\subsection{Any2Poster Agent}
\label{app:agent_details}

\ouralg{} follows a parse--chunk--analyze--plan--generate--compile pipeline.
The parser converts heterogeneous inputs into a shared \texttt{ParsedDocument}
schema. Long sections are split into chunks of roughly 800 tokens, with local
context windows retained for downstream reasoning. The global analysis stage
predicts the poster title, core message, section importance, candidate panels,
and source figures to preserve. The section-level analysis stage produces
panel-ready summaries, panel roles, layout modes, and candidate visual
suggestions.

The poster is rendered as HTML/CSS. Text is rendered directly in HTML to
preserve character-level fidelity, while visual assets are either extracted
from the source or generated from panel-specific prompts. The final poster is
compiled through a browser renderer and exported as PDF and PNG. A VLM-based
feedback loop optionally crops poster panels, diagnoses common visual issues,
and applies localized HTML/CSS edits.

\subsection{Visual Feedback Rules}
\label{app:feedback_rules}

The VLM feedback loop returns coarse diagnostic labels for each panel:
\textit{overflow}, \textit{too blank}, \textit{visual too small},
\textit{visual too large}, or \textit{good}. These labels are mapped to
deterministic repair actions. Overflow reduces font scale and may remove the
least important final bullet. Blank panels increase visual prominence or
spacing. Undersized figures receive larger flex allocation, while oversized
figures are scaled down. The poster is then re-rendered. This local repair
strategy avoids regenerating the full poster and helps preserve already
correct content.

\section{Ablation Study}
\label{app:ablation}

\begin{table}[h]
\centering
\small
\setlength{\tabcolsep}{6pt}
\caption{Ablation study of \ouralg{} under the PaperQuiz-style setting with 50
verbatim and 50 interpretive questions. Higher is better for all metrics.}
\label{tab:ablation}
\begin{tabular}{lcccc}
\toprule
\textbf{Variant} &
\textbf{Verbatim} &
\textbf{Interpretive} &
\textbf{Overall Acc.} &
\textbf{Density-Aug.} \\
\midrule
Full \ouralg{} &
\textbf{60.80} & \textbf{84.00} & \textbf{72.58} & \textbf{145.16} \\
w/o unified input parsing &
31.00 & 83.00 & 58.00 & 114.00 \\
w/o visual feedback refinement &
42.00 & 69.00 & 55.50 & 111.00 \\
\bottomrule
\end{tabular}
\end{table}

Removing unified input parsing causes the largest drop in verbatim accuracy,
indicating that structured source recovery is especially important for
fine-grained facts. Removing visual feedback refinement also reduces overall
accuracy, showing that visual quality control affects whether downstream
readers can recover information from the final poster.

\section{Efficiency}
\label{app:efficiency}\label{sec:efficiency}

We report token consumption to characterize the computational cost of different
poster-generation pipelines. As shown in Table~\ref{tab:token_efficiency},
\ouralg{}-4o uses 138.34K tokens per run on average, mainly from LLM calls.
This is higher than PosterAgent-4o and the direct 4o-HTML baseline, but lower
than PPTAgent-4o and OWL-4o. The direct 4o-HTML baseline has the lowest token
consumption, but it reflects a simpler direct-generation setting rather than a
full agentic poster-generation pipeline. Overall, \ouralg{}-4o has moderate
computational cost while supporting heterogeneous input parsing, poster
planning, generation, compilation, and feedback refinement.

\begin{table}[h]
\centering
\small
\setlength{\tabcolsep}{8pt}
\caption{Average token consumption per run across poster-generation methods.
Lower is better.}
\label{tab:token_efficiency}
\begin{tabular}{lc}
\toprule
\textbf{Method} & \textbf{Avg. Tokens} \\
\midrule
4o-HTML & 20.67K \\
PosterAgent-Qwen & 47.55K \\
PosterAgent-4o & 101.10K \\
\ouralg{}-4o & 138.34K \\
PPTAgent-4o & 255.73K \\
OWL-4o & 361.10K \\
\bottomrule
\end{tabular}
\end{table}

\section{Additional Qualitative Examples}
\label{app:qualitative}\label{sec:qualitative}

We include additional qualitative examples to illustrate the diversity of
\ourbench{} across modalities and domains. These examples are intended to show
how poster-generation systems handle heterogeneous source structures, such as
technical papers, webpages, notebooks, slide decks, and narrative prose.

\IfFileExists{assets/multipleexamples.pdf}{%
\begin{figure*}[h]
    \centering
    \includegraphics[width=\textwidth]{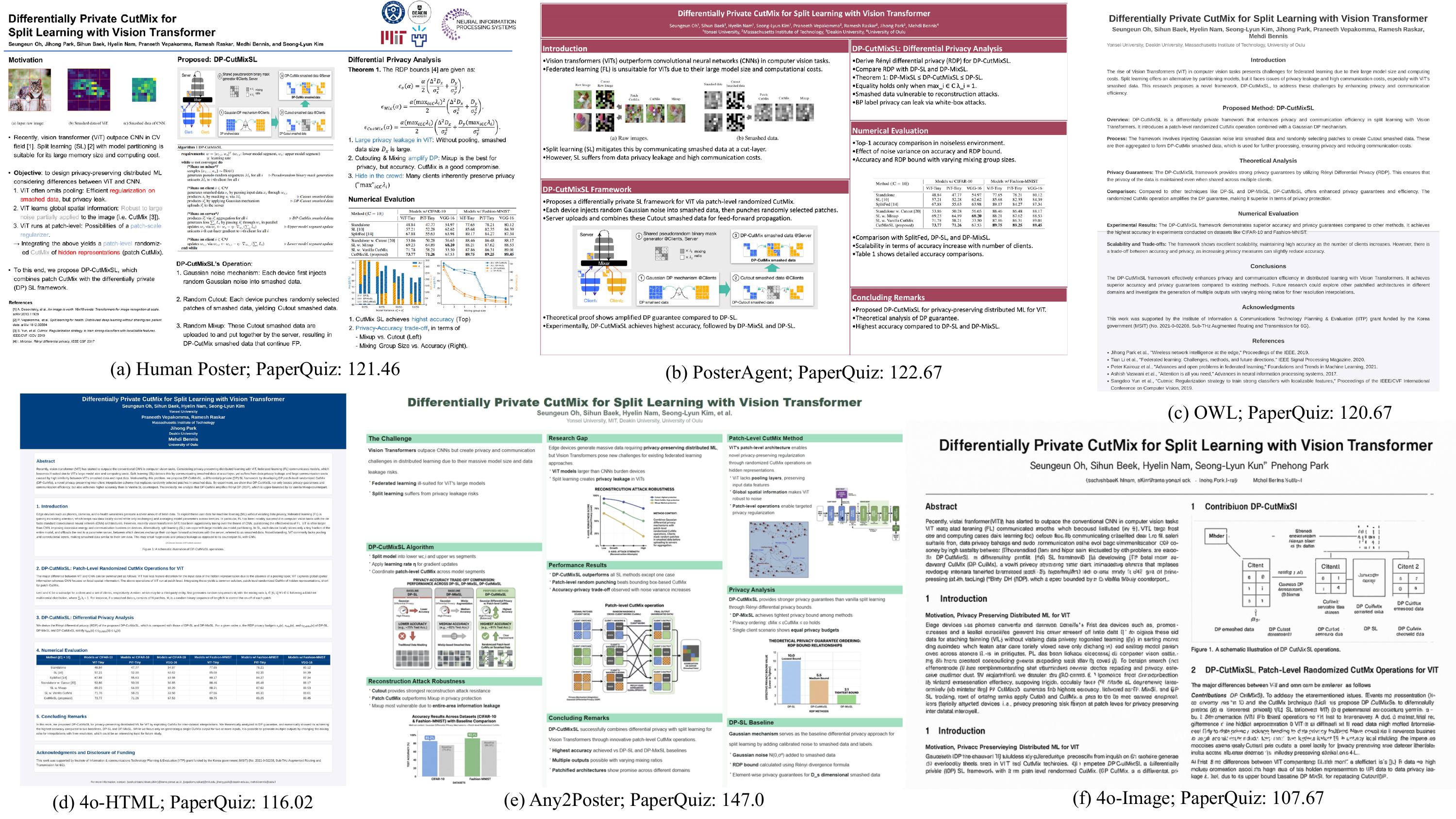}
    \caption{Qualitative examples comparing generated posters from different
    systems on the same source.}
    \label{fig:a2p_examples_app}
\end{figure*}
}{}

\IfFileExists{assets/a2p-additional-examples.png}{%
\begin{figure*}[h]
    \centering
    \includegraphics[width=\textwidth]{assets/a2p-additional-examples.png}
    \caption{Additional qualitative examples from \ourbench{} across input
    modalities and content domains.}
    \label{fig:a2p_additional_examples}
\end{figure*}
}{}

\section{Limitations}
\label{app:limitations}\label{sec:limitations}

\ourbench{} broadens poster-generation evaluation beyond paper-only inputs, but
it is not exhaustive. The benchmark currently focuses on English-language
sources and a finite set of eight input modalities and five content domains.
Future versions could expand to more languages, more specialized domains, and
additional source types such as interactive webpages, spreadsheets, or
multi-file project repositories.

The evaluation protocol also has limitations. BenchQuiz measures whether
information can be recovered from a generated poster by VLM readers, but VLM
performance may vary across models and may not perfectly match human reader
behavior. VLM-as-judge scores provide scalable visual-quality assessment, but
they may reflect model-specific preferences and should not be interpreted as a
complete replacement for human evaluation. To reduce bias, we randomize
answer-option order and allow \texttt{NA} responses, but residual model bias may
remain.

\ouralg{} is a reference system rather than a final solution to any-source
poster generation. Its performance depends on parser quality, LLM planning
quality, visual-generation quality, and the reliability of the VLM feedback
loop. Difficult inputs such as heavily formatted slide decks, noisy videos,
documents with complex tables, or sources requiring precise visual reproduction
remain challenging. In addition, the use of large multimodal models introduces
computational cost and may limit accessibility for some users.

\section{Responsible Release}
\label{app:responsible_release}\label{sec:responsible}

The benchmark is intended for research on evaluating and improving poster
generation systems. It should not be used to claim factual correctness beyond
the provided source content, nor should it be used as a standalone measure of
general design ability. For release, we document source provenance, modality,
content domain, processing status, and evaluation files. When source licensing
does not permit redistribution, we provide metadata or pointers rather than
redistributing the original content.

We also prepare Croissant-compatible metadata and Responsible AI metadata
describing intended uses, limitations, source provenance, licensing
considerations when available, and recommended evaluation practices. Generated
posters may summarize or visually transform copyrighted source materials, so
users should ensure that downstream usage complies with the source license and
applicable policies.

\end{document}